\documentclass[conference]{IEEEtran}
\IEEEoverridecommandlockouts

\usepackage{cite}
\usepackage{amsmath,amssymb,amsfonts}
\usepackage{algorithmic}
\usepackage{float}
\usepackage{graphicx}
\usepackage{textcomp}
\usepackage{xcolor}
\usepackage[hidelinks]{hyperref}
\usepackage[nameinlink,capitalise,noabbrev]{cleveref}
\crefname{figure}{Fig.}{Figs.}
\Crefname{figure}{Fig.}{Figs.}
\crefname{table}{Table}{Tables}
\Crefname{table}{Table}{Tables}
\crefname{equation}{Eq.}{Eqs.}
\Crefname{equation}{Eq.}{Eqs.}

\usepackage{doi}
\def\BibTeX{{\rm B\kern-.05em{\sc i\kern-.025em b}\kern-.08em
    T\kern-.1667em\lower.7ex\hbox{E}\kern-.125emX}}

\usepackage{fancyhdr}
\fancypagestyle{firstpage}{
    \fancyhf{} 
    \fancyhead[L]{Accepted at 2025 28th International Conference on Computer and Information Technology (ICCIT)\\
    19--21 December 2025, Cox’s Bazar, Bangladesh}

}

\begin{document}

\title{Balancing Accuracy and Efficiency: CNN Fusion Models for Diabetic Retinopathy Screening}

\author{
\IEEEauthorblockN{Md Rafid Islam}
\IEEEauthorblockA{\textit{Department of Electrical and Computer Engineering} \\
\textit{North South University} \\
Dhaka, Bangladesh \\
md.islam.241@northsouth.edu}
\and
\IEEEauthorblockN{Rafsan Jany}
\IEEEauthorblockA{\textit{Digital Health Research Division} \\
\textit{Korea Institute of Oriental Medicine} \\
Daejeon, South Korea \\
rafsanjany@kiom.re.kr}
\and
\IEEEauthorblockN{Akib Ahmed}
\IEEEauthorblockA{\textit{Department of Computer Science} \\
\textit{American International University--Bangladesh} \\
Dhaka, Bangladesh \\
akib.ahmed@aiub.edu}
\and
\IEEEauthorblockN{Mohammad Ashrafuzzaman Khan}
\IEEEauthorblockA{\textit{Department of Electrical and Computer Engineering} \\
\textit{North South University} \\
Dhaka, Bangladesh \\
mohammad.khan02@northsouth.edu}
}

\maketitle
\thispagestyle{firstpage}  
\begin{abstract}

Diabetic retinopathy (DR) remains a leading cause of preventable blindness, yet large-scale screening is constrained by limited specialist availability and variable image quality across devices and populations. This work investigates whether feature-level fusion of complementary convolutional neural network (CNN) backbones can deliver accurate and efficient binary DR screening on globally sourced fundus images. Using 11,156 images pooled from five public datasets (APTOS, EyePACS, IDRiD, Messidor, and ODIR), we frame DR detection as a binary classification task and compare three pretrained models (ResNet50, EfficientNet-B0, and DenseNet121) against pairwise and tri-fusion variants. Across five independent runs, fusion consistently outperforms single backbones. The EfficientNet-B0 + DenseNet121 (Eff+Den) fusion model achieves the best overall mean performance (accuracy: 82.89\%) with balanced class-wise F1-scores for normal (83.60\%) and diabetic (82.60\%) cases. While the tri-fusion is competitive, it incurs a substantially higher computational cost. Inference profiling highlights a practical trade-off: EfficientNet-B0 is the fastest (approximately 1.16 ms/image at batch size 1000), whereas the Eff+Den fusion offers a favorable accuracy--latency balance. These findings indicate that lightweight feature fusion can enhance generalization across heterogeneous datasets, supporting scalable binary DR screening workflows where both accuracy and throughput are critical.

\end{abstract}


\section{Introduction}

Diabetic retinopathy (DR) occurs when diabetes damages the blood vessels in the retina and is a potentially vision-threatening complication of diabetes mellitus. This damage, if left untreated, may result in blindness \cite{b1}. DR is a leading cause of blindness worldwide, emphasizing the importance of early detection to preserve vision \cite{b2}.

Traditional methods of DR detection include fundus photography and ophthalmoscopy, which are labor-intensive and require a long time for both image acquisition and interpretation \cite{b3}. Diagnosis relies heavily on the expertise and judgment of ophthalmologists, which can vary between practitioners, and the requirement of highly trained specialists and specialized facilities makes it inaccessible in resource-limited settings \cite{b4}. AI is ushering in a revolutionary change in DR detection by addressing significant limitations of traditional screening methods \cite{b5}. These technologies facilitate rapid, accurate, and scalable diagnosis and may serve as a valuable tool for ophthalmologists to reinforce their judgment \cite{b6}.

However, the performance of a single automated model can be limited by its architectural biases and sensitivity to specific image characteristics. Depending on a single decision source may be insufficient in clinical screening, where false negatives (missed cases) carry severe consequences. Fusion models, which combine the complementary strengths of multiple architectures, offer an effective solution. Even a slight improvement can be clinically meaningful in large-scale screening programs, where a small 1--2\% increase in accuracy or sensitivity means many additional patients being correctly identified.

We experimented with pre-trained deep learning models like ResNet50, EfficientNetB0, and DenseNet121 and developed fusion models that synergize the strengths of these three architectures, enhancing accuracy and reliability. Several recent studies have initiated the integration of multiple DR datasets, but comprehensive evaluations encompassing up to five distinct sources with careful benchmarking remain underexplored. The main contributions of this work can be summarized as follows:
\begin{itemize}
    \item We trained and evaluated on a combined dataset (5 sources) for global generalization.  
    
    \item We systematically evaluate three widely used CNN backbones (ResNet50, EfficientNet-B0, and DenseNet121) for the binary classification of DR using fundus images.
    
    \item We propose fusion-based architectures, including both pairwise and tri-fusion models, that integrate complementary feature representations from multiple backbones.
    
    \item We evaluate efficiency using parameter count, model size, and inference time to reveal accuracy–cost trade-offs.
\end{itemize}

\section{Related Work}

Deep learning (DL) for diabetic retinopathy (DR) detection has evolved rapidly over the past decade. The advent of end-to-end DL marked an important turning point, enabling models to directly extract features from fundus images without manual intervention \cite{b7}. Reviews of this field generally trace the progression through three overlapping stages: lesion detection, severity grading, and segmentation, illustrating the broader shift from traditional machine learning to data-driven DL methods \cite{b8}.

One of the first influential applications came from Gulshan \textit{et al.} (2016), who used the Inception-v3 model to detect referable DR in the large EyePACS dataset \cite{b9}. Their work achieved impressive sensitivity but also highlighted challenges such as class imbalance. Meanwhile, segmentation emerged as an equally important line of research, since identifying microaneurysms, exudates, and vessel abnormalities is crucial for clinical interpretability. U-Net-based models have been particularly successful; for example, Li \textit{et al.} (2019) reported Dice scores exceeding 95\% for vessel and lesion segmentation on DRIVE and STARE datasets \cite{b10}. Alyoubi \textit{et al.} (2020) highlighted the promise of hybrid CNN--RNN designs for predicting disease progression over time \cite{b8}. Tsiknakis \textit{et al.} (2021) showed how GAN-based data augmentation improved performance on imbalanced datasets such as IDRiD \cite{b11}.

Alqudah \textit{et al.} (2024) demonstrated that lightweight ensembles of EfficientNet and custom CNNs achieved 98.5\% accuracy on APTOS while remaining practical for mobile deployment \cite{b6}. Similarly, Sharma \textit{et al.} (2024) proposed parameter-efficient frameworks that reduced model size by 70\% yet achieved 97\% AUC on EyePACS, showing feasibility in low-resource environments \cite{b12}. Gong and Pu (2025) advanced deep learning into predictive modeling by integrating convolutional neural networks (CNNs) with long short-term memory networks (LSTMs) to predict diabetic retinopathy (DR) using clinical datasets, achieving a precision of 92\% and surpassing traditional statistical benchmarks \cite{b13}. Furthermore, systematic reviews such as Alshammari \textit{et al.} (2025) highlighted a growing adoption of vision transformers, which demonstrate superior generalization across diverse populations and imaging conditions \cite{b14}.

While ensemble and fusion techniques have been explored, comprehensive benchmarking of fusion architectures combining multiple modern backbones (e.g., ResNet50, EfficientNetB0, DenseNet121) on heterogenous multi-source datasets for binary DR screening remains limited, and our work directly addresses this gap. 

\section{Methodology}

\begin{figure*}[t]
    \centering
    \includegraphics[width=0.9\textwidth]{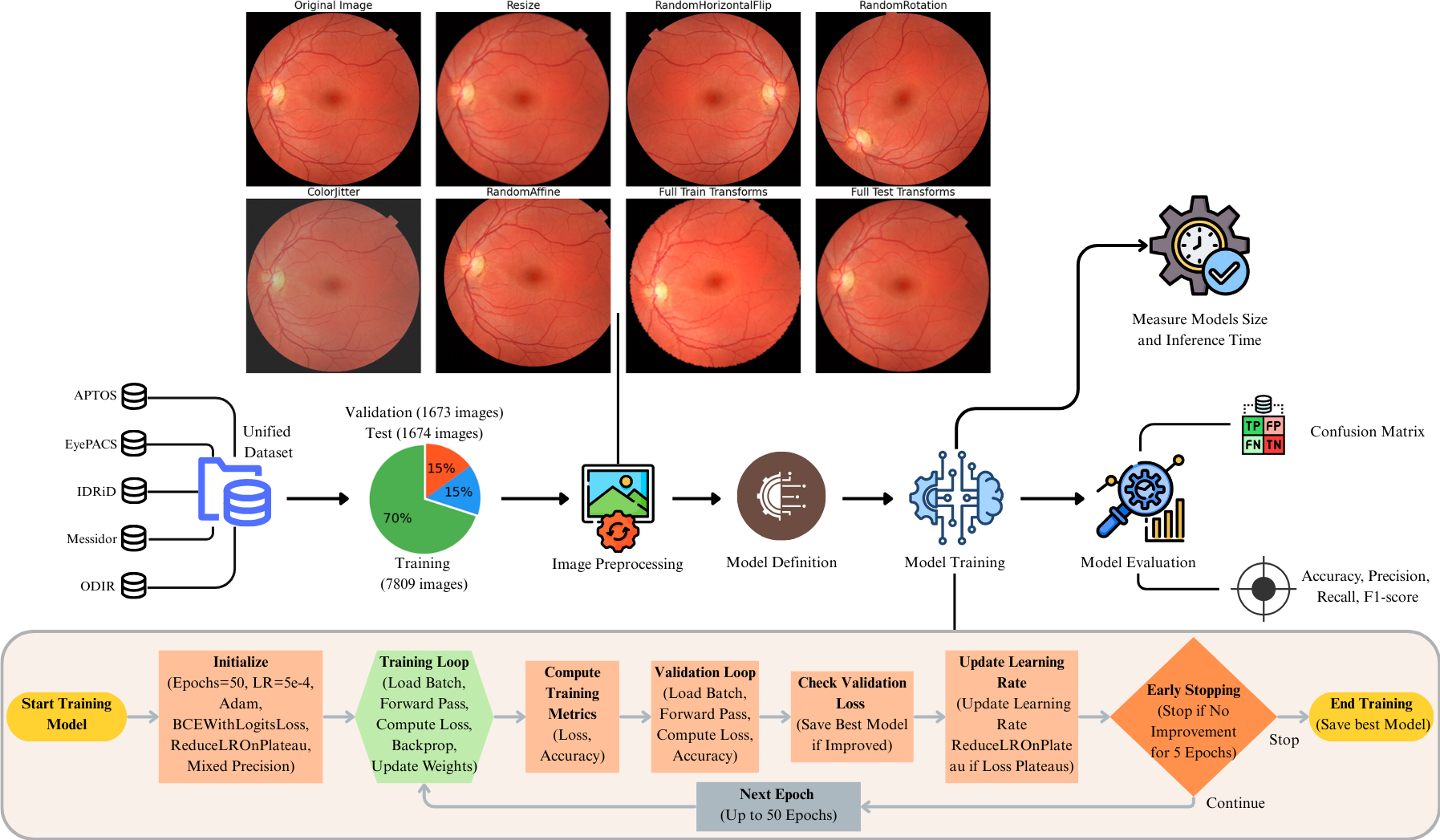}
    \caption{Comprehensive DR detection workflow: preprocessing, training, and evaluation.}
    \label{fig:workflow}
\end{figure*}

First, we collected data from five different datasets and merged them into a single dataset. After preprocessing, the data was split into training, validation, and testing sets. Next, we experimented with three different pretrained models and evaluated the results. After that, with the combination of these three models, we trained and evaluated fusion models and compared the performance of the individual base models with the fusion models. The entire workflow of training and evaluating a model is presented in Fig.~\ref{fig:workflow}.

\subsection{Dataset Description}

The combined dataset comprises a total of 11,156 retinal fundus images collected from five public sources, organized into two binary classes: \textit{Diabetic} (images showing signs of DR) and \textit{Normal} (images without DR). The contributing datasets are APTOS, EyePACS, IDRiD, Messidor, and ODIR \cite{b15,b16,b17,b18,b19}. The distribution of images across these sources is presented in Table~\ref{tab:dataset_distribution}.

\begin{table}[htbp]
\caption{Distribution of Images in the Combined Dataset}
\label{tab:dataset_distribution}
\centering
\setlength{\tabcolsep}{4pt}
\renewcommand{\arraystretch}{1.2}
\begin{tabular}{|p{10mm}|p{27mm}|p{8mm}|c|c|c|}
\hline
\textbf{Source} & \textbf{Description} & \textbf{Region} & \textbf{Diabetic} & \textbf{Normal} & \textbf{Total} \\
\hline
APTOS & High-quality publicly curated images & India & 1,857 & 1,805 & 3,662 \\
\hline
EyePACS & Large, diverse, variable quality & USA & 588 & 490 & 1,078 \\
\hline
IDRiD & Annotated for DR and lesions & India & 279 & 168 & 447 \\
\hline
Messidor & Diverse resolutions, dilation status varied & France & 731 & 1,014 & 1,745 \\
\hline
ODIR & Multi-disease, high diversity & China & 2,123 & 2,101 & 4,224 \\
\hline

\end{tabular}
\end{table}

\subsection{Data Preprocessing and Augmentation}

All images were resized to $224 \times 224$ pixels to ensure compatibility with pretrained deep learning architectures. The dataset was divided into training (70\%), validation (15\%), and test (15\%) sets using stratified sampling to preserve class balance across the splits. To improve model generalization and mitigate overfitting, several data augmentation techniques were applied to the training set, including random horizontal flipping, random rotation ($\pm 15^{\circ}$), color jitter (brightness and contrast adjustment), and random affine transformations with slight translation and scaling. For the validation and test sets, only resizing and normalization were applied. All images were normalized using the standard ImageNet mean and standard deviation.

\subsection{Base Models}

\subsubsection{ResNet50}
A strong CNN backbone with deep hierarchical feature extraction. The model introduces skip connections (identity shortcuts) to address vanishing gradients in deep networks and uses 1$\times$1, 3$\times$3, and 1$\times$1 convolutional layers to reduce parameters while maintaining performance.

\subsubsection{EfficientNetB0}
Enhances feature efficiency while preserving accuracy and uniformly scales depth, width, and resolution using a fixed coefficient. It consists of nine MBConv blocks with varying kernel sizes (3$\times$3 to 5$\times$5) and expansion ratios.

\subsubsection{DenseNet121}
Known for its densely connected architecture, which enhances feature reuse and improves the flow of information and gradients throughout the network. Each layer receives inputs from all preceding layers and passes its own feature maps to all subsequent layers within the same dense block.

\subsection{Fusion Model}

\subsubsection{Fusion Strategy}
First, \textit{feature-level fusion} extracts high-level features from the penultimate layer of each base model, concatenates them, and feeds the combined representation into a final classifier. Second, \textit{logit fusion} aggregates the final predictions (logits) of individual models through averaging or weighted combination. Finally, \textit{hybrid fusion} combines both feature-level and logit-level information to leverage the strengths of each approach.

\subsubsection{Fusion Architecture}
The base networks operate as parallel feature extractors, and their outputs are fused to form a unified representation for final classification. Let an input fundus image be denoted as:
\begin{equation}
\mathbf{X} \in \mathbb{R}^{3 \times H \times W}, \quad H = W = 224
\end{equation}
where $\mathbf{X}$ has three RGB channels. The input is passed through $n$ pretrained CNN backbones ($n=2$ for pairwise fusion, $n=3$ for tri-fusion), each truncated before the final classification layer to act as a high-level feature encoder:
\begin{equation}
\mathbf{f}_i = M_i(\mathbf{X}) \in \mathbb{R}^{d_i}, \quad i = 1,2,\dots,n
\end{equation}
where $M_i(\cdot)$ denotes the feature extraction function of the $i$-th pretrained model, and $d_i$ is the dimension of its extracted feature vector.  

These features are concatenated to form a composite feature representation:
\begin{equation}
\mathbf{f}_{\text{concat}} = [\mathbf{f}_1; \mathbf{f}_2; \dots; \mathbf{f}_n] \in \mathbb{R}^{\sum_{i=1}^n d_i}
\end{equation}

A fully connected linear layer is applied to reduce the concatenated vector to a lower-dimensional representation:
\begin{equation}
\mathbf{f}_{\text{dr}} = \mathbf{W}_{\text{dr}} \mathbf{f}_{\text{concat}} + \mathbf{b}_{\text{dr}} \in \mathbb{R}^{d}
\end{equation}
where $\mathbf{W}_{\text{dr}} \in \mathbb{R}^{d \times \sum d_i}$, $\mathbf{b}_{\text{dr}} \in \mathbb{R}^{d}$, and $d$ is a tunable hyperparameter.  

This reduced vector $\mathbf{f}_{\text{dr}}$ is passed through a two-layer classification head consisting of a dense layer with ReLU activation, dropout (rate = 0.5), and a final fully connected layer producing a logit score:
\begin{align}
\mathbf{h}_1 &= \text{ReLU}(\mathbf{W}_1 \mathbf{f}_{\text{dr}} + \mathbf{b}_1) \in \mathbb{R}^{512} \\
\mathbf{h}_{1}^{\text{drop}} &= \text{Dropout}(\mathbf{h}_1, p=0.5) \\
y &= \mathbf{W}_2 \mathbf{h}_{1}^{\text{drop}} + \mathbf{b}_2 \in \mathbb{R}
\end{align}
where $y$ is the logit score corresponding to the presence of diabetic retinopathy. The final probability is obtained via sigmoid activation:
\begin{equation}
P(y=1 \mid \mathbf{X}) = \sigma(y) = \frac{1}{1 + \exp(-y)}.
\end{equation}

The model is trained using the Binary Cross-Entropy with Logits Loss (BCEWithLogitsLoss), which combines a sigmoid layer and binary cross-entropy in a numerically stable form:
\begin{equation}
\mathcal{L}(y, \hat{y}) = - y \log \sigma(\hat{y}) - (1-y)\log(1-\sigma(\hat{y}))
\end{equation}
where $y \in \{0,1\}$ is the ground truth label and $\hat{y}$ is the predicted logit.

\begin{figure*}[t]
    \centering
    \includegraphics[width=0.9\textwidth]{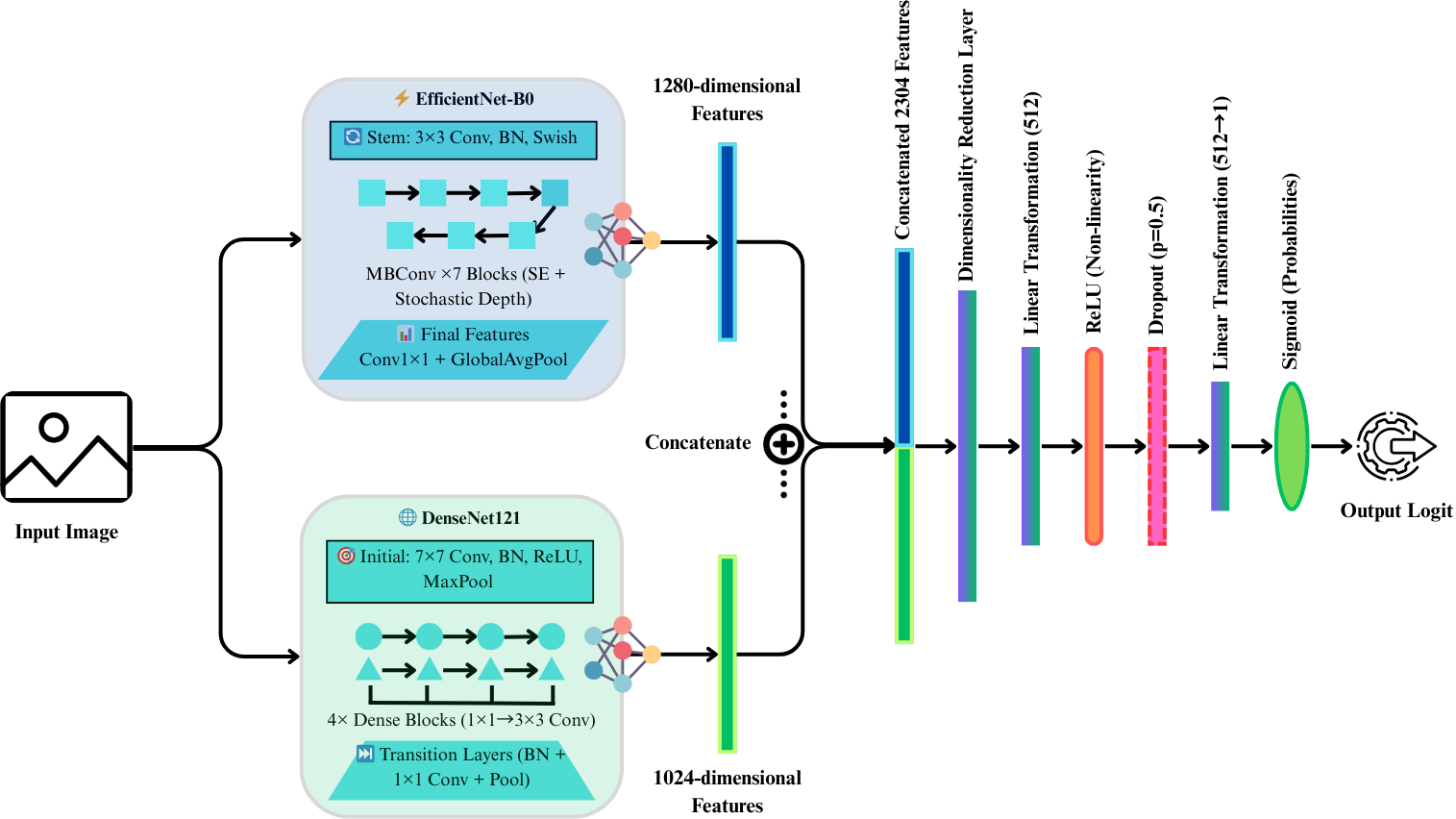}
    \caption{Architecture of the best-performing pairwise fusion model (EfficientNet-B0 + DenseNet121).}
    \label{fig:fusion_architecture}
\end{figure*}

We designed three pairwise fusion models: ResNet50 + EfficientNet-B0 (Res+Eff), ResNet50 + DenseNet121 (Res+Den), and EfficientNet-B0 + DenseNet121 (Eff+Den), and a tri-fusion model (Res+Eff+Den). Although multiple fusion architectures were tested, only the diagram of the best-performing pairwise fusion model (EfficientNet-B0 + DenseNet121) is presented as a representative example, as shown in Fig.~\ref{fig:fusion_architecture}.

\subsection{Training Procedure}

Each model was trained using a supervised learning framework with \texttt{BCEWithLogitsLoss}, suitable for binary classification tasks. The Adam optimizer was used with a learning rate of $0.001$ and a weight decay of $1 \times 10^{-4}$ for $\ell_2$ regularization and a \texttt{ReduceLROnPlateau} scheduler to reduce the learning rate by a factor of $0.1$ if the validation loss plateaued for three consecutive epochs. If no improvement in validation loss was observed over five consecutive epochs, training was terminated via early stopping. All training metrics (loss and accuracy for both training and validation phases) were logged using TensorBoard, enabling real-time monitoring of model performance and detection of potential overfitting. The best-performing model (based on minimum validation loss) was saved and subsequently used for final evaluation on the independent test set.

\subsection{Evaluation Metrics and Testing}

After training, the best-performing model weights (based on validation loss) were loaded for evaluation on the held-out test set. The following standard evaluation metrics for binary classification were employed to assess model performance:

\subsubsection{Accuracy}

Accuracy reflects the ratio of correctly classified images (normal and diabetic) to the total number of test samples:
\begin{equation}
\text{Accuracy} = \frac{TP + TN}{TP + TN + FP + FN}
\end{equation}

\subsubsection{Precision}
Precision measures the proportion of images predicted as positive that were actually positive:
\begin{equation}
\text{Precision} = \frac{TP}{TP + FP}
\end{equation}

\subsubsection{Recall}
Recall measures the proportion of actual positive images that were correctly identified by the model:
\begin{equation}
\text{Recall} = \frac{TP}{TP + FN}
\end{equation}

\subsubsection{F1-Score}
The F1-score is the harmonic mean of precision and recall, providing a balanced measure of performance:
\begin{equation}
\text{F1-Score} = \frac{2 \times \text{Precision} \times \text{Recall}}{\text{Precision} + \text{Recall}}
\end{equation}

\subsection{Measuring Inference Time}

Inference latency was evaluated across batch sizes of 1, 10, 100, and 1000. All experiments were conducted using PyTorch on a workstation equipped with an Intel Core i7-10870H CPU @ 2.20~GHz, 16~GB RAM, and an NVIDIA GeForce RTX~3060 Laptop GPU with 6~GB VRAM. Test images were loaded from the designated directory, preprocessed into tensors, and grouped into the specified batch sizes for measurement. To stabilize GPU clocks, multiple warm-up iterations were performed prior to measurement. For each batch size, inference was run three times, and both the total batch inference time and the average per-image inference time were recorded, and preprocessing time was excluded.

\section{Experimental Results and Analysis}

\subsection{Overall Results}

\begin{figure}[H]
    \centering
    \includegraphics[width=\linewidth]{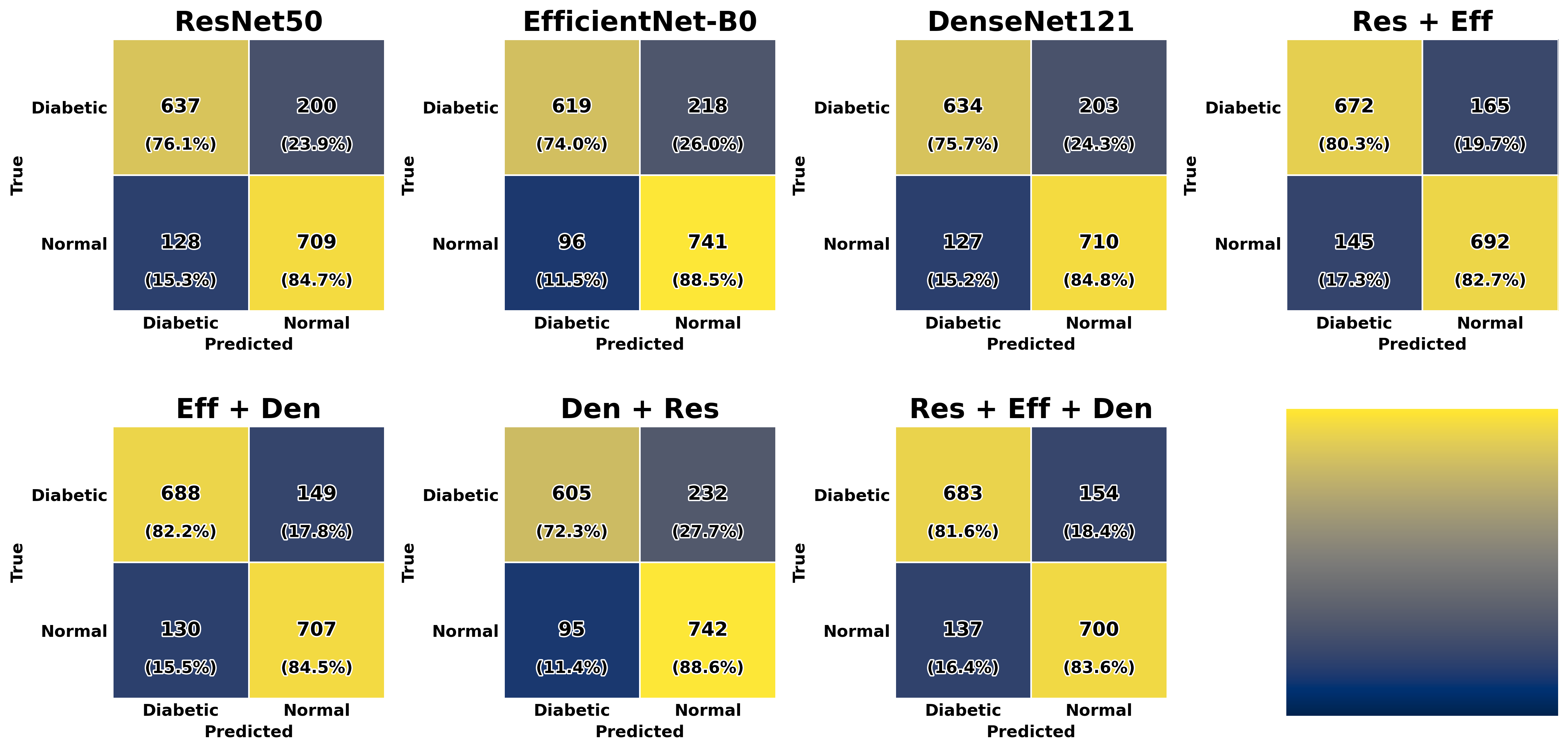}
    \caption{Confusion matrices for all models from the best-performing run.}
    \label{fig:confusion_matrix}
\end{figure}

Each model was trained and evaluated five times. The mean and standard deviation for each evaluation metric were calculated to ensure a statistically robust performance assessment and to mitigate the influence of individual fortunate or unfortunate runs. The aggregated results are reported in Table~\ref{tab:overall_results}. For all analyses, the diabetic retinopathy (DR) class is treated as the positive class. Additionally, the confusion matrices from the best-performing run of each model are presented in Fig.~\ref{fig:confusion_matrix} which provide further insight into the distribution of true and false predictions.

\begin{table*}[t]
\caption{Overall Performance of Base and Fusion Models (Mean $\pm$ Standard Deviation Over 5 Runs)}
\label{tab:overall_results}
\centering
\setlength{\tabcolsep}{4pt}
\renewcommand{\arraystretch}{1.1}
\begin{tabular}{|l|c|c|c|c|c|c|c|}
\hline
\textbf{Metric} & \textbf{ResNet50} & \textbf{EfficientNet-B0} & \textbf{DenseNet121} & \textbf{Res+Eff} & \textbf{Eff+Den} & \textbf{Den+Res} & \textbf{Res+Eff+Den} \\
\hline
Test Loss & 0.41 $\pm$ 0.012 & 0.38 $\pm$ 0.009 & 0.39 $\pm$ 0.010 & 0.38 $\pm$ 0.005 & 0.37 $\pm$ 0.009 & 0.39 $\pm$ 0.013 & 0.37 $\pm$ 0.012 \\
Accuracy (\%) & 78.31 $\pm$ 1.33 & 80.61 $\pm$ 0.82 & 80.00 $\pm$ 0.32 & 80.27 $\pm$ 0.81 & 82.89 $\pm$ 0.25 & 79.43 $\pm$ 0.86 & 82.33 $\pm$ 0.16 \\
\hline
\multicolumn{8}{|l|}{\textit{Normal Class}} \\
\hline
Precision & 76.20 $\pm$ 2.56 & 77.33 $\pm$ 1.25 & 78.40 $\pm$ 1.02 & 81.40 $\pm$ 2.87 & 81.20 $\pm$ 1.47 & 75.60 $\pm$ 1.36 & 82.00 $\pm$ 1.90 \\
Recall & 82.60 $\pm$ 4.18 & 86.67 $\pm$ 1.97 & 83.40 $\pm$ 1.96 & 79.00 $\pm$ 5.76 & 85.60 $\pm$ 3.07 & 86.80 $\pm$ 2.99 & 83.20 $\pm$ 3.54 \\
F1-Score & 79.20 $\pm$ 1.47 & 81.50 $\pm$ 0.96 & 80.40 $\pm$ 0.49 & 80.00 $\pm$ 1.67 & 83.60 $\pm$ 0.49 & 80.80 $\pm$ 0.98 & 82.20 $\pm$ 0.75 \\
\hline
\multicolumn{8}{|l|}{\textit{Diabetic Class}} \\
\hline
Precision & 81.00 $\pm$ 2.83 & 84.67 $\pm$ 1.60 & 82.00 $\pm$ 1.26 & 79.80 $\pm$ 3.12 & 85.00 $\pm$ 2.28 & 84.60 $\pm$ 2.65 & 83.00 $\pm$ 2.28 \\
Recall & 74.20 $\pm$ 4.35 & 74.83 $\pm$ 2.19 & 76.60 $\pm$ 2.06 & 81.80 $\pm$ 4.53 & 80.00 $\pm$ 2.76 & 72.00 $\pm$ 3.16 & 81.00 $\pm$ 3.25 \\
F1-Score & 77.60 $\pm$ 1.74 & 79.33 $\pm$ 0.94 & 79.20 $\pm$ 0.75 & 80.40 $\pm$ 1.02 & 82.60 $\pm$ 0.49 & 78.00 $\pm$ 1.41 & 82.00 $\pm$ 0.63 \\
\hline
\end{tabular}
\end{table*}

\subsection{Ablation and Comparative Evaluation}

\begin{figure}[htbp]
    \centering
    \includegraphics[width=\linewidth]{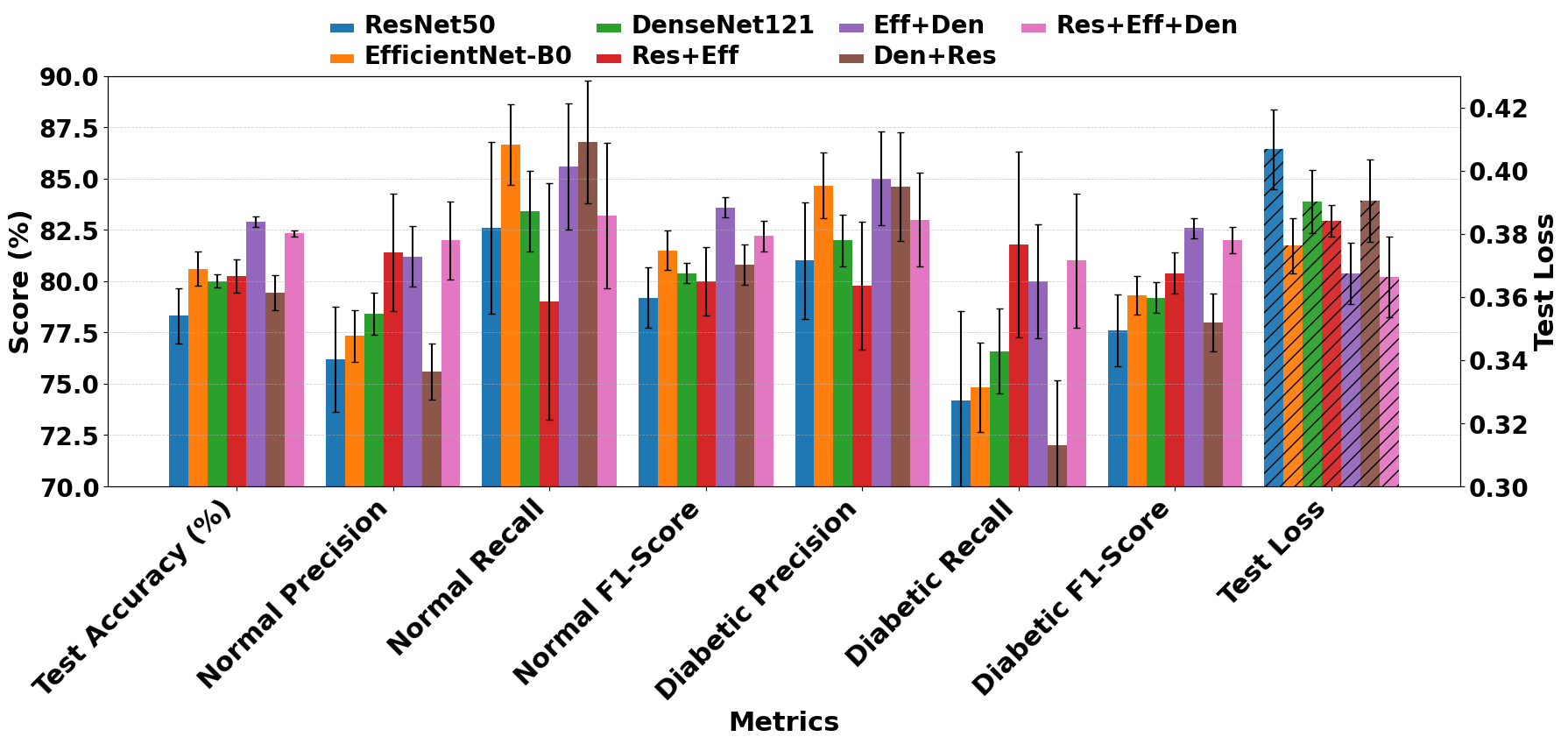}
    \caption{Comparative visualization of evaluation metrics across all models.}
    \label{fig:comp_mat}
\end{figure}

Ablation studies and comparative analysis were conducted to assess the contribution of each backbone and the efficacy of different fusion models. A comparative summary of model performance is shown in Fig.~\ref{fig:comp_mat}. The results reveal three key patterns. First, single-model performance varied by architecture: EfficientNet-B0 performed best (80.61\% accuracy), followed by DenseNet121 (80.00\%), while ResNet50 lagged behind (78.31\%). The class-wise F1-scores also showed a similar ranking, implying that ResNet50 may be less effective for our heterogeneous, multi-source fundus data.

Second, feature-level fusion consistently outperforms single models, but the gains depend on the complementary nature of the backbones. The Eff+Den fusion achieved the highest overall accuracy (82.89\%) and the best-balanced F1-scores for both normal (83.60\%) and diabetic (82.60\%) classes. Den+Res provided little improvement and reduced normal class precision, which indicates limited complementarity. The Res+Eff fusion offered a recall boost for diabetic cases but did not surpass EfficientNet-B0's precision.

Third, the tri-fusion model provides a recall-oriented alternative at a computational premium. While its accuracy (82.33\%) is slightly lower than Eff+Den, it achieved the highest recall for diabetic cases (81.00\%), which can be useful for screening where missing true positives is costly. However, this came with a ~3× increase in parameters compared to Eff+Den (Table~\ref{tab:model_size}), which makes it less suitable for resource-constrained deployment. Overall, Eff+Den stands out as the most accurate and efficient model.  

\subsection{Model Size and Inference Time}

\begin{figure}[t]
    \centering
    \includegraphics[width=\linewidth]{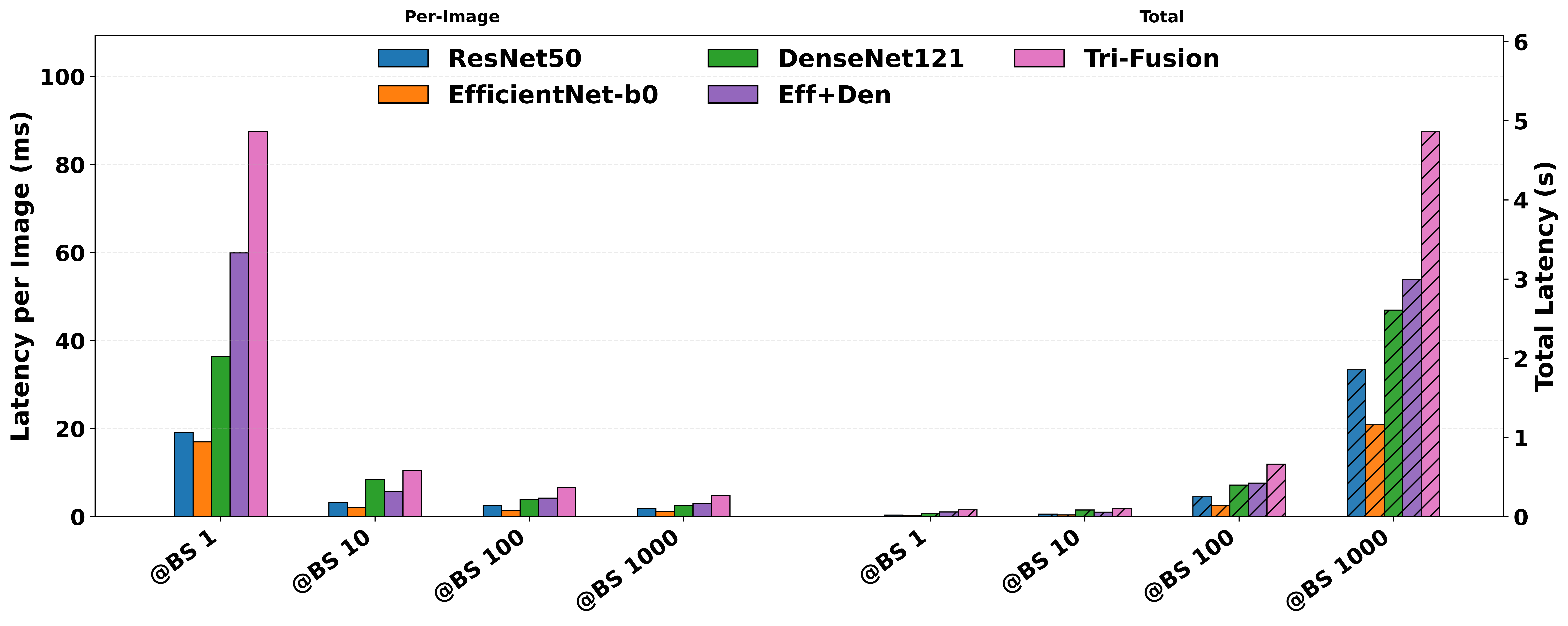}
    \caption{Inference latency across batch sizes for base models and the Eff+Den fusion model: (a) total batch inference time (seconds); (b) per-image inference time (milliseconds).}
    \label{fig:inference}
\end{figure}

The practical deployability of deep learning models is heavily influenced by their computational footprint. This subsection reports parameter counts, storage needs, and inference times for the base models, tri-fusion, and the best Eff+Den fusion model. The size of each model, quantified by the number of trainable parameters (in millions) and its corresponding storage footprint on disk (in megabytes), is summarized in Table~\ref{tab:model_size}. Total inference time (seconds) and per-image inference time (milliseconds) across batch sizes of 1, 10, 100, and 1000 are illustrated in Fig.~\ref{fig:inference}. The tri-fusion model, due to its increased complexity and size, exhibits the highest total and per-image inference times among all evaluated architectures. EfficientNet-B0 is the fastest model across all batch sizes, achieving a per-image latency as low as 1.16~ms at a batch size of 1000. The Eff+Den fusion provides a favorable balance between classification performance, model size, and inference latency. Notably, the per-image inference time consistently decreases as batch size increases, indicating improved throughput in high-volume screening environments.

\begin{table}[t]
\centering
\caption{Model Size in Terms of Trainable Parameters and Disk Size}
\label{tab:model_size}
\begin{tabular}{|l|c|c|}
\hline
\textbf{Model} & \textbf{Parameters (M)} & \textbf{Size on Disk (MB)} \\
\hline
EfficientNet-B0 & 4.66  & 18.0  \\
DenseNet121     & 7.48  & 29.1  \\
ResNet50        & 24.56 & 93.9  \\
Eff+Den Fusion  & 15.29 & 59.2  \\
Tri-Fusion      & 45.06 & 173.0 \\
\hline
\end{tabular}
\end{table}

\section{Conclusion}

This study set out to develop a dependable, deployable DR screening pipeline by (i) training and evaluating widely used CNN backbones on a pooled, multi-regional fundus dataset and (ii) testing whether feature/logit fusion can offer measurable gains over single models. Among all candidates, the Eff+Den model delivered the strongest overall performance (highest mean accuracy and balanced classwise F1) while maintaining a modest computational footprint. The tri-fusion variant further reduced error in some settings but at a substantial cost in parameters and latency. Together with the confusion-matrix analysis and inference benchmarks, these results suggest two viable deployment paths: EfficientNet-B0 when ultra-low latency is paramount, and Eff+Den when slightly higher latency is acceptable in exchange for stronger and more stable accuracy. Our evaluation also underscores the value of training on diverse sources for improved robustness to variation in acquisition protocols, resolutions, and populations, an essential property for real-world screening. Future work will extend the framework to multi-grade DR severity, explore self-supervised backbones, investigate domain adaptation and fairness across underrepresented cohorts, and integrate model-aware triage (fast first-pass models backed by fusion-based confirmers). By rigorously balancing accuracy with computational efficiency, this work contributes a step toward practical DR screening in resource-limited clinical environments.

\end{document}